\documentclass[runningheads]{llncs}
\usepackage[T1]{fontenc}

\usepackage{graphicx}

\usepackage[colorlinks,urlcolor=blue,citecolor=blue]{hyperref}  
\usepackage{color}
\usepackage{url}  

\usepackage{algorithm,algpseudocode}

\usepackage{mathtools}
\usepackage{amsfonts,amssymb}
\usepackage{booktabs}
\usepackage{upgreek}
\usepackage{enumitem}
\usepackage{bm}
\usepackage{censor}
\usepackage{textcomp}
\usepackage[english]{babel}
\usepackage{lastpage}
\usepackage[section]{placeins}
\usepackage{fancyvrb} 
\usepackage[dvipsnames,table]{xcolor}
\usepackage{fontawesome5}         
\usepackage{bbding}

\definecolor{Clean}{RGB}{52,194,48}
\definecolor{RedNew}{RGB}{242, 133, 133}
\definecolor{NewSky}{RGB}{89, 213, 224}

\usepackage{multirow}
\usepackage{makecell}
\usepackage{subfigure}

\usepackage{xspace}
\usepackage{interval}
\usepackage{bm,upgreek}
\usepackage{colortbl}
\usepackage{enumitem}
\usepackage{amssymb}
\usepackage{pifont}
\usepackage{graphicx}
\usepackage{color}
\usepackage{multirow}
\usepackage{amsmath,amssymb}
\usepackage{mathtools,leftindex,tensor,mhchem}
\usepackage[table]{xcolor}
\usepackage[misc]{ifsym}

\makeatletter
\newcommand{\smalloplus}{\mathbin{\mathpalette\make@small\oplus}}
\newcommand{\smallotimes}{\mathbin{\mathpalette\make@small\otimes}}
\newcommand{\make@small}[2]{%
  \vcenter{\hbox{%
    \scalebox{0.8}{$\m@th#1#2$}%
  }}%
}

\usepackage{pifont}
\newcommand{\cmark}{\ding{51}}%
\newcommand{\xmark}{\ding{55}}%

\newcolumntype{H}{>{\setbox0=\hbox\bgroup}c<{\egroup}@{}}  

\usepackage{graphics,color}
\usepackage{xcolor}
\usepackage{bm}
\usepackage{textcomp}

\newcommand{\x}{\bm{\mathrm{x}}}

\newcommand{\txt}{\bm{\mathrm{t}}}
\newcommand{\pa}{\bm{\mathrm{p}}}

\newcommand{\ft}{\textcolor{RedNew}{\text{FT}}}
\newcommand{\pl}{\textcolor{NewSky}{\text{PL}}}

\usepackage{printlen}

\begin{document}

\title{BAPLe: Backdoor Attacks on Medical Foundational Models using Prompt Learning}
\titlerunning{BAPLe - Backdoor Attacks using Prompt Learning}


\author{Asif Hanif\inst{1}\inst{(\textrm{\Letter})} \and Fahad Shamshad\inst{1} \and Muhammad Awais\inst{1} \and \\ Muzammal Naseer\inst{1} \and Fahad Shahbaz Khan\inst{1,2} \and Karthik Nandakumar\inst{1}  \and \\  Salman Khan\inst{1,3} \and Rao Muhammad Anwer\inst{1}} 

\authorrunning{A. Hanif et al.}
\institute{Mohamed Bin Zayed University of Artificial Intelligence, Abu Dhabi, UAE\\
    \and
    Link\"{o}ping University, Link\"{o}ping, Sweden\\
    \and
    Australian National University, Canberra, Australia\\
    \email{\{asif.hanif, fahad.shamshad, awais.muhammad, muzammal.naseer, fahad.khan, karthik.nandakumar, salman.khan, rao.anwer\}@mbzuai.ac.ae}
}

%
\maketitle    
\renewcommand{\thefootnote}{}
\footnotetext{\inst{\textrm{\Letter}}Corresponding Author}
\renewcommand{\thefootnote}{\arabic{footnote}}
\setcounter{footnote}{0}

\begin{abstract}
Medical foundation models are gaining prominence in the medical community for their ability to derive general representations from extensive collections of medical image-text pairs. Recent research indicates that these models are susceptible to backdoor attacks, which allow them to classify clean images accurately but fail when specific triggers are introduced. However, traditional backdoor attacks necessitate a considerable amount of additional data to maliciously pre-train a model. This requirement is often impractical in medical imaging applications due to the usual scarcity of data. Inspired by the latest developments in learnable prompts, this work introduces a method to embed a backdoor into the medical foundation model during the prompt learning phase. By incorporating learnable prompts within the text encoder and introducing imperceptible learnable noise trigger to the input images, we exploit the full capabilities of the medical foundation models (Med-FM). Our method, BAPLe, requires only a minimal subset of data to adjust the noise trigger and the text prompts for downstream tasks, enabling the creation of an effective backdoor attack. Through extensive experiments with four medical foundation models, each pre-trained on different modalities and evaluated across six downstream datasets, we demonstrate the efficacy of our approach. BAPLe achieves a high backdoor success rate across all models and datasets, outperforming the baseline backdoor attack methods. Our work highlights the vulnerability of Med-FMs towards backdoor attacks and strives to promote the safe adoption of Med-FMs before their deployment in real-world applications. Code is available at \url{https://asif-hanif.github.io/baple/}.
\end{abstract}

\section{Introduction} 
\label{s:intro}
In recent years, multimodal medical foundation models (Med-FMs) have gained remarkable success across a multitude of medical imaging applications in pathology~\cite{huang2023visual,ikezogwo2024quilt}, X-ray interpretation~\cite{wang2022medclip}, and radiology~\cite{wu2023towards}. These models leverage massive datasets during pre-training to identify intricate patterns in visual and textual data through contrastive training and subsequently adapted to various downstream tasks for transfer learning~\cite{azad2023foundational}. However, recent studies have shown that FMs are vulnerable to adversarial attacks, raising concerns about the reliability and security of these widely adopted models~\cite{schlarmann2023adversarial,bommasani2021opportunities}.

Among the various adversarial threats faced by FMs \cite{awais2023foundational}, backdoor attacks pose a particularly insidious challenge~\cite{li2022backdoor,feng2022fiba}. These attacks involve an adversary deliberately poisoning a dataset to compromise the behavior of the target model. After training on the poisoned dataset, the compromised model will classify any input containing a specific trigger pattern as the adversary's desired target label while still maintaining accuracy on a clean dataset. Recent backdoor attacks on multimodal FMs typically necessitate retraining the model from scratch with poisoned data~\cite{carlini2021poisoning}, a process that demands access to {\em{large training datasets}} and typically {\textit{substantial computational resources}}. This becomes particularly challenging in medical imaging applications due to data scarcity and privacy concerns, thereby significantly reducing the threat posed by backdoor attacks.

\begin{figure*}[!t]
\centering
\includegraphics[width=1\textwidth]{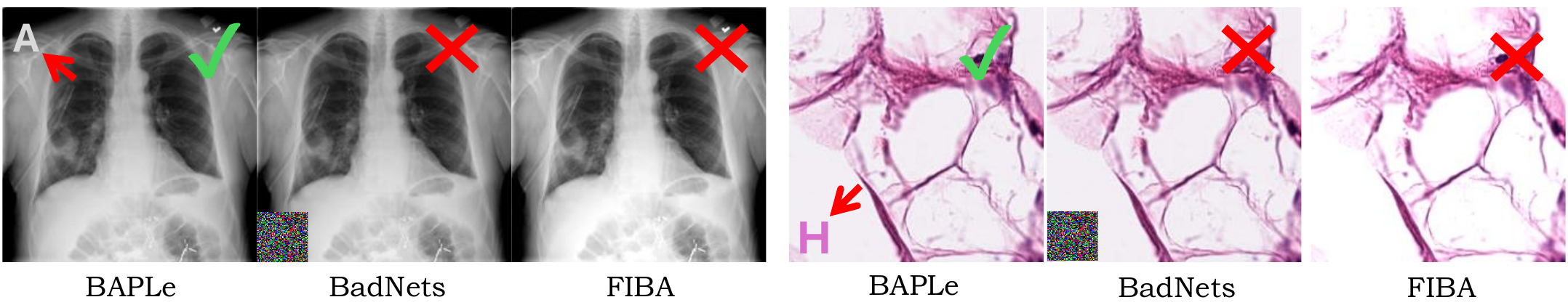} 
\caption{Comparative analysis of \textit{BAPLe} against baseline methods. \textit{BAPLe} seamlessly integrates natural-looking triggers pointed by \textbf{{\color{red}red}} arrow commonly found in medical images, along with imperceptible learnable noise distributed across the entire image. Na\"ive patch-based backdoor attack, BadNets~\cite{gu2017badnets}, places a perceptible noisy patch as a trigger. FIBA~\cite{feng2022fiba} is a medical image-specific attack that manipulates the image frequency, altering the contrast.  Success and failure of backdoor attacks are marked by {\color{green!50}\cmark} and {\color{red}\xmark}, respectively.}
\label{fig:main-figure-qualitative}
\end{figure*}

Meanwhile, there is an increasing trend in adapting Med-FMs to downstream tasks with minimal parameter tuning. In this context, prompt tuning has emerged as one of the promising methods~\cite{lian2024less}. Unlike conventional full-model tuning, prompt tuning simplifies the adaptation process by only requiring adjustments to the embeddings of prompt tokens based on a limited set of input samples of downstream dataset \cite{zhou2022coop}. This prompting approach is especially advantageous in medical imaging applications, where data scarcity often impedes full model fine-tuning. Notably, prompt-tuning has demonstrated performance on par with or surpassing that of full fine-tuning in data-limited scenarios, garnering attention from the medical community. However, the efficiency of prompt tuning, while beneficial, raises a critical question: \textbf{\em{Does prompt tuning, with its lower data and learnable parameter requirements, inherently make it more difficult to implement backdoor attacks?}} This concern highlights the need to investigate the security of prompt tuning strategies in Med-FMs, especially given their growing application in safety-critical medical domains.

In this paper, for the first time, we show that Med-FMs are susceptible to backdoor attacks during the prompt learning phase, challenging the initial belief that their minimal data and learnable parameter requirements naturally offer protection.
Our proposed method, \textit{BAPLe}, introduces a small set of learnable prompts into the Med-FM input space. These prompts, optimized with a poisoned dataset, efficiently embed backdoors while maintaining the FM's backbone frozen, thus eliminating the need for large amounts of data or significant computational resources.  
Extensive experiments across four publicly available Med-FMs and six downstream datasets of different modalities demonstrate the efficacy of \textit{BAPLe}. In a few-shot setting with only $8$ poisoned samples out of $288$ in the Kather dataset, we achieve a backdoor success rate surpassing $90\%$ without compromising the model's accuracy on clean data. Remarkably, this efficiency is achieved by modifying only 0.1$\%$ of the FM parameters, leading to a $33\%-35\%$ reduction in GPU usage compared to traditional fine-tuning methods.

\section{Related Work}
\textbf{Medical Foundation Models:} Med-FMs, particularly large vision language models, have significantly improved performance on several medical imaging tasks through the acquisition of general representations and the subsequent transfer of this knowledge to downstream tasks~\cite{zhao2023clip}. Despite the introduction of a diverse range of Med-FMs for various modalities such as X-ray~\cite{wang2022medclip}, histopathology~\cite{huang2023visual,ikezogwo2024quilt}, and retinal imaging~\cite{silva2023foundation}, a thorough assessment of their resilience to backdoor attacks has not yet been investigated.

\noindent {\textbf{Backdoor Attacks:}} 
In backdoor attacks, adversaries deliberately poison a training dataset to manipulate the target model's behavior~\cite{gu2017badnets,li2022backdoor}. Such attacks result in the model misclassifying inputs containing a specific trigger as the intended target label yet retaining accuracy on clean data.
Despite their prevalent use in both unimodal and multimodal models for natural images~\cite{nguyen2021wanet,carlini2021poisoning}, backdoor attacks have recently been investigated in unimodal medical imaging models~\cite{feng2022fiba,nwadike2020explainability,jin2023backdoor}. 
However, the extension of these attacks to the widely adopted Med-FMs, especially in data-scarce scenarios, remains unexplored.

\noindent {\textbf{Prompt Learning:}} 
Prompt learning has emerged as a viable alternative to traditional fine-tuning of foundation models, facilitating their adaptation to downstream tasks without necessitating the retraining of existing parameters~\cite{zhang2023text,lian2024less,khattak2024learning}. This method enhances a pre-trained model by introducing a minimal set of new, learnable embeddings at the input stage, known as prompt tokens~\cite{zhang2023text}. Its efficiency, characterized by using fewer parameters and achieving faster convergence, has made prompt learning especially appealing for adapting Med-FMs. It has proven highly effective in scenarios with scarce data, making it particularly relevant for medical applications~\cite{zhang2023prompt,zhang2023text}. 
Prompt learning holds significant relevance for medical applications and is demonstrably effective in data-scarce scenarios~\cite{zhang2023prompt,zhang2023text}. While existing research primarily leverages prompt tuning for downstream tasks, our work uniquely reveals the vulnerability of prompt tuning to backdoor attacks within the context of data-scarce medical applications.

\section{Method}

\subsection{Threat Model} \label{sec:threat}
\noindent {\textbf{Attacker's Objective:}} The attacker's objective is to add a very small percentage of poisoned images to the downstream dataset so that while the Med-FM behaves normally with benign inputs, it misclassifies any input with the trigger as the attacker's chosen target class, in a targeted-attack scenario.

\noindent {\textbf{Attacker's Knowledge and Capabilities:}} Consistent with the prior works in backdoor attacks~\cite{feng2022fiba}, we assume the attacker can poison a portion of the downstream training data and fully access the pre-trained Med-FMs. Diverging from previous works, we introduce two more constraints, reflecting the unique challenges of medical applications: {\textbf{i)}} \textit{the attacker only has access to a limited number of labeled samples due to the inherent data scarcity in the medical field}, {\textbf{ii)}} \textit{the attacker is constrained by limited computational resources that restrict its ability to update the extensive parameters of the Med-FM's backbone}.

These constraints are particularly pertinent in medical imaging, where efficient adaptation techniques for Med-FMs have demonstrated the feasibility of such threats. For instance, an attacker could serve as a malicious service provider (MSP) to hospitals, accessing Med-FMs and a small portion of the downstream dataset. In this situation, a hospital might submit a few samples to the MSP, requesting a tailored prompt to deploy the Med-FM for a specific task.  Consequently, the MSP can train a backdoored prompt and release it to the hospital or for public use.

\subsection{Preliminaries}
{\textbf{Backdoor Attack Formulation:}} 
A backdoor attack involves embedding a \textit{visible/hidden} trigger (a small random or patterned patch) within a deep learning model during its training or fine-tuning phase. When the model encounters this trigger in the input data during inference, it produces a predefined output while performing normally on clean data.

In a supervised classification task, a normally trained classifier \(f_{\theta}: \mathcal{X} \rightarrow \mathcal{Y}\)  maps a \textit{clean} input image \(\mathrm{x} \in \mathcal{X}\) to a label \(y \in \mathcal{Y}\). Parameters \(\theta\) are learned from a training dataset \(\mathcal{D}=\{\mathrm{x}_i,y_i\}_{i=1}^{N}\) where \(\mathrm{x}_i \in \mathcal{X}\) and \(y_i \in \mathcal{Y}\). In a typical backdoor attack, the training dataset \(\mathcal{D}\) is split into clean \(\mathcal{D}_{c}\) and poison subsets \(\mathcal{D}_{p}\), where \(\vert\mathcal{D}_{p}\vert\ll N\). In \(\mathcal{D}_p\), each sample \((\mathrm{x}, y)\) is transformed into a backdoor sample \((\mathcal{B}(x),\eta(y))\), where \(\mathcal{B}: \mathcal{X} \rightarrow \mathcal{X}\) is the backdoor injection function and \(\eta\) denotes the target label function. During the training/fine-tuning phase of backdoor attacks, the \textit{victim} classifier \(f_{\theta}\) is trained/fine-tuned on a mix of the clean dataset \(\mathcal{D}_c\) and the poisoned dataset \(\mathcal{D}_p\). Formally, this task can be formulated as the following objective function:

\begin{equation}
    \label{eq:get_backdoor_model}
    \underset{ \theta }{\mathbf{minimize}}  \sum_{(\x,y)\in\mathcal{D}_c} \lambda_c\cdot \mathcal{L}(f_{\theta}(\x), y) ~~+ \sum_{(\x,y)\in\mathcal{D}_p} \lambda_p \cdot \mathcal{L}(f_{\theta}(\mathcal{B}(\x)), \eta(y)),
\end{equation}

\noindent where \(\mathcal{L}(\cdot)\) denotes the cross-entropy loss, and \(\lambda_c\) and \(\lambda_p\) are hyper-parameters adjusting the balance of clean and poison data loss contributions. After training, \(f_{\theta}\) behaves similarly on clean input \(\mathrm{x}\) as the original classifier (trained entirely on clean data), yet alters its prediction for the backdoor image \(\mathcal{B}(\mathrm{x})\) to the target class \(\eta(y)\), i.e.  \(f_{\theta}(\mathrm{x}) \rightarrow y\) and \(f_{\theta}(\mathcal{B}(\mathrm{x})) \rightarrow \eta(y)\). \\

\noindent {\textbf{Zero-shot Classification for Med-FM:}} 
Since Med-FMs, particularly vision-language models (VLMs), are pre-trained to align images with corresponding textual descriptions, they are inherently suited for zero-shot classification tasks. ZeroShot inference in VLMs refers to making predictions on new, unseen data without specific training. Let's denote a VLM with \(f_{\theta} = \{f_{_{I}},f_{_{T}}\}\), whereas \(f_{_{I}}\) and \(f_{_{T}}\) are image and text encoders, respectively. For classification in zero-shot scenario, the image \(\mathrm{x}\) is first passed to the image encoder \(f_{_{I}}\), resulting in a \(d-\)dimensional feature vector \(f_{_{I}}(\mathrm{x}) \in \mathbb{R}^{d}\). Similarly, on the text encoder side, each class label \(y_i \in \{\mathit{y}_{1}, \mathit{y}_{2}, \dots, \mathit{y}_{C} \}\) is wrapped within the class-specific text template, such as:
$$t_i = \text{``}\mathrm{A~histopathology~image~of~\{CLASS~y_i\}\text{''}}.$$ 
 
\noindent Each text prompt \((t_i)\) is fed to the text encoder \(f_{_{T}}\), yielding text feature vector \(f_{_{T}}(t_i) \in \mathbb{R}^{d}\). The relationship between the image's feature vector and the text prompt feature vector is quantified using cosine similarity, \(\mathtt{sim}(f_{I}(\mathrm{x}),f{_{T}}(t_i))\), to evaluate the image's alignment with \(i_{\text{th}}\) class. Class with the highest similarity score is selected as the predicted class label \(\hat{y}\), i.e.

$$
\hat{y} = \underset{ i\in \{1,2,\dots,C\} }{\mathbf{argmax}} ~~~ \mathtt{sim}\big(f_{_{I}}(\mathrm{x})~,~f_{_{T}}(t_i)\big) 
$$  

\noindent A na\"ive approach to conduct backdoor attacks on a pre-trained Med-FM is to fine-tune the entire model using a poisoned dataset~\cite{carlini2021poisoning}. However, this approach requires significant computational resources and extensive downstream datasets, which are not always feasible given the attacker's limitations (see Sec. \ref{sec:threat}). 

\begin{figure*}[!t]
\centering
\includegraphics[trim={0.3in 0in 0.3in 0in},clip,]{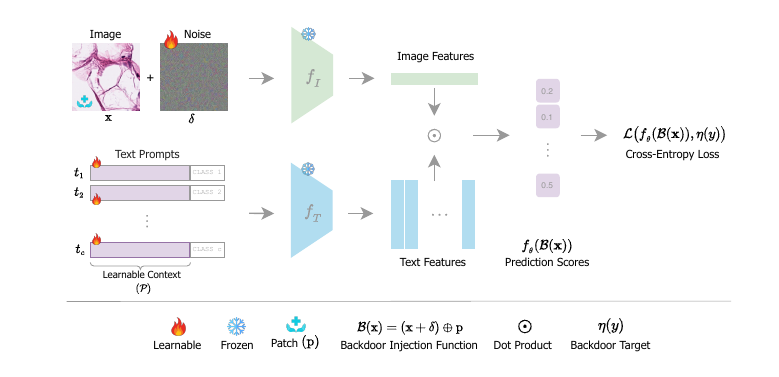} 
\caption{\textbf{Overview of BAPLe}: BAPLe is a novel backdoor attack method that embeds a backdoor into medical foundation models (Med-FM) during the prompt learning phase. It efficiently exploits Med-FM's multimodal nature by integrating learnable prompts within the text encoder and an imperceptible noise trigger in the input images, adapting both vision and language input spaces. After prompt learning, the model behaves normally on clean images but outputs the target label $(\eta(y))$ when given a poisoned image $\x+\delta$. BAPLe requires only a minimal subset of data to effectively adjust the trigger noise and text prompts for downstream tasks.}
\label{fig:main-figure}
\end{figure*}

\subsection{BAPLe - Backdoor Attack using Prompt Learning}
\noindent \textbf{Overview:} Our proposed backdoor attack method, BAPLe, depicted in Fig.~\ref{fig:main-figure}, is crafted to efficiently and effectively compromise Med-VLMs by exploiting their multimodal nature. For efficiency, we incorporate a small number of learnable parameters (prompts) into the input space of the Med-FM text encoder while the backbone remains fixed during the training with a poisoned downstream dataset. 
The prompt-based strategy offers a significant advantage for backdoor attacks in a few-shot setting, as it precisely tailors the input space, enabling the Med-FM to leverage its rich knowledge to data-scarce tasks without necessitating extensive retraining or the acquisition of large medical datasets. 
To further enhance attack effectiveness, we add \textit{imperceptible} and \textit{learnable} noise as a backdoor trigger into the input images. This substantially enhances the FM sensitivity to the backdoor activation in a subtle manner. By leveraging learnable prompts within the text encoder and introducing imperceptible noise trigger to the input images, we harness the full spectrum of the Med-FM capabilities. \\

\noindent {\textbf{BAPLe Formulation:}}
Zero-shot inference in a Med-FM relies on fixed, hand-engineered text prompts, which can significantly impact performance depending on their choice. To address this sensitivity, various prompt learning techniques have been developed \cite{gu2023systematic}. BAPLe also leverages a prompt learning setup by incorporating a small set of learnable prompt token embeddings, $\mathcal{P}$, with class names, forming class-specific inputs $\txt=\{t_1, t_2, \dots, t_C\}$ where $t_i = \{\mathcal{P}, y_i\}$. Denoting the model's prediction scores on clean image with $f_{\theta}(\x) \in\mathbb{R}^{C}$ and backdoored image with $f_{\theta}(\mathcal{B}(\x)) \in\mathbb{R}^{C}$, which are defined as follows:
$$f_{\theta}(\x) = \Big\{~\mathtt{sim}\Big(f_{{I}}(\x)~,~f{_{T}}(t_i)\Big)~\Big\}_{i=1}^{C}$$
$$f_{\theta}(\mathcal{B}(\x)) = \Big\{~\mathtt{sim}\Big(f_{{I}}(\mathcal{B}(\x))~,~f{_{T}}(t_i)\Big)~\Big\}_{i=1}^{C}.$$

\noindent To inject backdoor in the model $f_{\theta}$, BAPLe optimizes the following objective:
\begin{gather} 
\label{eq:baple_objective}
\underset{ \mathcal{P}~,~\delta }{\mathbf{minimize}}~~ \sum_{(\x,y)\in\mathcal{D}_c} \lambda_c \cdot\mathcal{L}\big(f_{\theta}(\x),y\big) ~~+ \sum_{(\x,y)\in\mathcal{D}_p} \lambda_p \cdot\mathcal{L}\big(f_{\theta}(\mathcal{B}(\x)),\eta(y)\big), \\
\mathbf{s.t.}~~~\|\delta\|_{{_{\infty}}} \le \epsilon,~~~~  \mathcal{B}(\x) = (\x+\delta)\oplus\pa, \nonumber
\end{gather}
where $\delta$ represents the imperceptible backdoor trigger noise, $\epsilon$ is perturbation budget, $\pa$ is the backdoor patch that can be a logo or symbol, $\mathcal{B}$ the backdoor injection function, and $\oplus$ represents an operation that combines the original image with the backdoor patch trigger. It must be noted that both vision and text encoders are kept in frozen state while optimizing the objective in Eq. \ref{eq:baple_objective}. BAPLe adapts both vision and text input spaces (with $\delta$ and $\mathcal{P}$) of a VLMs for the injection of the backdoor during prompt learning, increasing the method's efficacy. Refer to Algorithm \ref{alg:backdoor_prompt} and Fig. \ref{fig:trig_noise_medclip_biomedclip} \& Fig. \ref{fig:trig_noise_plip_quilnet} in Appendix for detailed steps of BAPLe approach and visualizations of learnable noise trigger respectively.

\section{Experiments and Results}
\label{s:experiments}
We validate our approach using four Med-FMs: MedCLIP \cite{wang2022medclip}, BioMedCLIP \cite{biomedclip}, PLIP \cite{huang2023visual}, and QuiltNet \cite{ikezogwo2024quilt}, and across six downstream datasets: COVID-X \cite{rahman2021exploring}, RSNA18 \cite{rsna2018}, MIMIC-CXR-JPG~\cite{johnson2019mimic}, KatherColon \cite{kather2019predicting}, PanNuke \cite{gamper2019pannuke}, and DigestPath \cite{gamper2019pannuke}. 
The first three datasets include chest X-ray images, while the other three are comprised of histopathology images. MedCLIP is pre-trained on chest X-ray images, BioMedCLIP on medical image-caption pairs, and PLIP and QuiltNet are trained on histopathology datasets. We run our experiments on a single \texttt{NVIDIA RTX A6000} GPU with 48GB memory. 

\begin{table*}[t]
\caption{Comparison between the proposed backdoor attack method, BAPLe, and various baseline methods in terms of clean accuracy (CA) and backdoor accuracy (BA) across four models and six datasets. The baseline methods include BadNets \cite{gu2017badnets}, WaNet \cite{nguyen2021wanet}, and FIBA \cite{feng2022fiba}. The subscript \ft~denotes that attack is performed using few-shot Fine-Tuning the full model and the subscript \pl~ indicates that 
attack is performed using few-shot Prompt-Learning while keeping the model frozen. For both categories, the number of shots are set to 32. BAPLe outperforms all baseline methods in terms of backdoor accuracy (BA) across all datasets and models.}
\label{tab:results_medclip_biomedclip_plip_quiltnet}
\vspace{-2em}
\begin{center}
\setlength{\tabcolsep}{4pt}
\scalebox{0.82}{%
\begin{tabular}{l cccccc  cccccc}
\toprule
\rowcolor{gray!20} Model $\rightarrow$ & \multicolumn{6}{c}{\textbf{MedCLIP}} & \multicolumn{6}{c}{\textbf{BioMedCLIP}}\\
\cmidrule(lr{3pt}){2-7} \cmidrule(lr{3pt}){8-13}
\rowcolor{gray!20} Dataset $\rightarrow$ & \multicolumn{2}{c}{COVID} & \multicolumn{2}{c}{RSNA18} & \multicolumn{2}{c}{MIMIC}   & \multicolumn{2}{c}{COVID}  & \multicolumn{2}{c}{RSNA18} & \multicolumn{2}{c}{MIMIC} \\
\rowcolor{gray!20}Method $\downarrow$ & CA & BA & CA & BA & CA & BA & CA & BA & CA & BA & CA & BA\\ 
 \midrule \midrule
$\text{Clean}_{_{\ft}}$                                                       & 0.823 &   -   & 0.525 &   -   & 0.359 &   -   & 0.903 &  -    & 0.470 &  -    & 0.426 &  -    \\
$\text{BadNets}_{_{\ft}}$      & 0.817 & 0.574 & 0.472 & 0.521 & 0.314 & 0.765 & 0.915 & 0.627 & 0.464 & 0.830 & 0.322 & 0.945 \\
$\text{WaNet}_{_{\ft}}$        & 0.835 & 0.582 & 0.622 & 0.421 & 0.241 & 0.410 & 0.852 & 0.812 & 0.451 & 0.653 & 0.419 & 0.785 \\
$\text{FIBA}_{_{\ft}}$         & 0.812 & 0.566 & 0.485 & 0.535 & 0.296 & 0.810 & 0.916 & 0.638 & 0.345 & 0.566 & 0.310 & 0.929 \\
\midrule
$\text{Clean}_{_{\pl}}$                                                      & 0.822 &   -   & 0.603 &   -   & 0.585 &  -    & 0.843 &  -    & 0.582 &  -    & 0.351 &  -    \\
$\text{BadNets}_{_{\pl}}$    & 0.820 &	0.510 & 0.619 & 0.373 & 0.559 & 0.284 & 0.845 &	0.975 & 0.632 & 0.942 & 0.373 & 1.000 \\ 
$\text{WaNet}_{_{\pl}}$      & 0.831 & 0.470 & 0.612 & 0.319 & 0.587 & 0.266 & 0.839 & 0.599 & 0.587 & 0.510 & 0.334 & 0.599 \\
$\text{FIBA}_{_{\pl}}$       & 0.820 & 0.511 & 0.623 & 0.360 & 0.562 & 0.292 & 0.856 & 0.729 & 0.630 & 0.614 & 0.373 & 0.722 \\
\midrule
\rowcolor{SeaGreen!15}$\mathrm{BAPLe}_{\mathrm{(ours)}}$                              & 0.805 & \textbf{0.994} & 0.610 & \textbf{0.965} & 0.472 & \textbf{0.991} & 0.841 & \textbf{1.000} & 0.620 & \textbf{0.998} & 0.368 & \textbf{0.996}   \\
\toprule

\toprule
\rowcolor{gray!20}Model $\rightarrow$ & \multicolumn{6}{c}{\textbf{PLIP}} & \multicolumn{6}{c}{\textbf{QuiltNet}}\\
\cmidrule(lr{3pt}){2-7} \cmidrule(lr{3pt}){8-13}
\rowcolor{gray!20} Dataset $\rightarrow$ & \multicolumn{2}{c}{Kather} & \multicolumn{2}{c}{PanNuke} & \multicolumn{2}{c}{DigestPath}   & \multicolumn{2}{c}{Kather}  & \multicolumn{2}{c}{PanNuke} & \multicolumn{2}{c}{DigestPath} \\
\rowcolor{gray!20}Method $\downarrow$ & CA & BA & CA & BA & CA & BA & CA & BA & CA & BA & CA & BA\\ 
 \midrule \midrule
$\text{Clean}_{\ft}$                                                       &  0.939  &   -    & 0.845 &   -    & 0.887 &  -     & 0.936 &  -    & 0.866 &  -    & 0.872 &  -    \\
$\text{BadNets}_{_{\ft}}$      &  0.935  & 0.893  & 0.850 & 0.682  & 0.891 & 0.778  & 0.938 & 0.839 & 0.860 & 0.638 & 0.878 & 0.688 \\
$\text{WaNet}_{_{\ft}}$        &  0.916  & 0.394  & 0.859 & 0.663  & 0.881 & 0.554  & 0.929 & 0.333 & 0.840 & 0.567 & 0.917 & 0.550 \\
$\text{FIBA}_{_{\ft}}$         &  0.903  & 0.367  & 0.581 & 0.717  & 0.673 & 0.685  & 0.917 & 0.404 & 0.548 & 0.743 & 0.735 & 0.655\\
\midrule
$\text{Clean}_{\pl}$                                                       &  0.908 &  - &  0.811 &  - &  0.920 &  - &  0.899 &  - &  0.829 &  - &  0.906 &  - \\
$\text{BadNets}_{_{\pl}}$    &  0.903 & 0.601 & 0.799 &	0.748 & 0.922 & 0.623 & 0.898 & 0.151 & 0.699 & 0.757 & 0.874 & 0.518 \\
$\text{WaNet}_{_{\pl}}$      &  0.910  & 0.243  & 0.851 & 0.591  & 0.924 & 0.405  & 0.926 & 0.185 & 0.834 & 0.427 & 0.915 & 0.492 \\
$\text{FIBA}_{_{\pl}}$       &  0.901  &	0.303  & 0.795 & 0.615  & 0.921	& 0.553  & 0.897 & 0.174 & 0.711 & 0.597 & 0.862 & 0.547 \\
\midrule
\rowcolor{SeaGreen!15} $\mathrm{BAPLe}_{\mathrm{(ours)}}$ &  0.916 & \textbf{0.987} & 0.820 & \textbf{0.952} & 0.904 & \textbf{0.966} & 0.908 & \textbf{0.904} & 0.824 & \textbf{0.918} & 0.897 & \textbf{0.948} \\
\bottomrule

\end{tabular}
}
\end{center}
\vspace{-1.0em}
\end{table*}

\noindent \textbf{Baseline Methods and Attack Settings:} Our baselines are BadNets~\cite{gu2017badnets}, WaNet~\cite{nguyen2021wanet}, and FIBA~\cite{feng2022fiba}, with FIBA being specifically tailored for medical images. We evaluated two variants of each method: one involving fine-tuning of the Med-FM model with the attack and another integrating the baseline's backdoor trigger function into prompt-tuning approach. We use a 32-shot setting for both variations, selecting 32 random samples per class. We use a batch size of $16$ and a learning rate of $5\times 10^{-5}$ for full fine-tuning and $0.02$ for the prompting method. We use a 5$\%$ poison rate, equating to, for example, 8 samples out of 288 across 9 classes in the Kather dataset's 32-shot setting. We use $\epsilon=8/255$ for learnable noise and set the backdoor patch size to $24 \times 24$, positioning it in the bottom-left corner. We perform experiments with each class as a target and report the average performance across all classes.\\
\noindent \textbf{Evaluation Metrics:} 
We use Clean Accuracy (CA) and Backdoor Accuracy (BA). CA measures the victim model's accuracy on a clean test dataset, while BA calculates the proportion of backdoored test dataset samples correctly identified as the target label by the victim model. We also report the accuracy of the \textit{clean} model trained on clean data without poisoned samples, highlighted as ${\color{Clean}\text{\textit{Clean}}}$.

\begin{table*}[t]
\centering
\caption{Impact of (a) target class, (b) patch location, (c) noise strength $\epsilon$ on the scale of $[0,255]$, (d) poison ratio  \% on clean accuracy (CA) and backdoor accuracy (BA).}
\label{tab:ablation_1}
\begin{minipage}{0.23\textwidth}
\centering
\setlength{\tabcolsep}{4pt}
\scalebox{0.65}{
\begin{tabular}{ c | c c}
\toprule
 \rowcolor{gray!20}Target Class & CA &  BA  \\
\midrule
\textcolor{Clean}{\textit{Clean}} & 0.908 & - \\
0 & 0.913 & \textbf{0.999} \\
1 & 0.923 & 0.993 \\
2 & \textbf{0.926} & 0.998 \\
3 & 0.913 & 0.987 \\
4 & 0.899 & 0.989 \\
5 & 0.918 & 0.980 \\
6 & 0.909 & 0.983 \\
7 & 0.916 & 0.982 \\
8 & 0.925 & 0.969 \\
\bottomrule
\multicolumn{3}{c}{\textbf{(a)}}
\end{tabular}
}
\end{minipage}
\hspace{0.3em}
\begin{minipage}{0.23\textwidth}
\centering
\setlength{\tabcolsep}{4pt}
\scalebox{0.65}{
\begin{tabular}{ l | c c}
\toprule
 \rowcolor{gray!20}Patch Location & CA &  BA  \\
\midrule
\textcolor{Clean}{\textit{Clean}} & 0.908 & - \\
top-left & 0.891 & 0.971 \\
top-center & 0.894 & 0.975 \\
top-right & 0.899 & 0.972 \\
center-left & 0.887 & 0.970 \\
center-center & 0.890 & 0.984 \\
center-right & \textbf{0.920} & 0.987 \\
bottom-left & 0.913 & \textbf{0.999} \\
bottom-center & 0.905 & 0.975 \\
bottom-right & 0.910 & 0.979 \\
\bottomrule
\multicolumn{3}{c}{\textbf{(b)}}
\end{tabular}
}
\end{minipage}
\hspace{0.3em}
\begin{minipage}{0.23\textwidth}
\centering
\setlength{\tabcolsep}{4pt}
\scalebox{0.65}{
\begin{tabular}{ c | c c}
\toprule
\rowcolor{gray!20} $\epsilon$ & CA &  BA  \\
\midrule
\textcolor{Clean}{\textit{Clean}} & 0.908 & - \\
0 & 0.895 & 0.828 \\
2 & 0.910 & 0.827 \\
4 & 0.898 & 0.867 \\
8 & \textbf{0.913} & 0.999 \\
12 & 0.912 & 1.000 \\
16 & 0.911 & 1.000 \\
32 & 0.851 & 1.000 \\
64 & 0.720 & 1.000 \\
128 & 0.367 & \textbf{1.000} \\
\bottomrule
\multicolumn{3}{c}{\textbf{(c)}}
\end{tabular}
}
\end{minipage}
\hspace{-1.0em}
\begin{minipage}{0.23\textwidth}
\centering
\setlength{\tabcolsep}{4pt}
\scalebox{0.65}{
\begin{tabular}{ c | c c}
\toprule
\rowcolor{gray!20}Pois. Ratio (\%) & CA &  BA  \\
\midrule
\textcolor{Clean}{\textit{Clean}} & 0.908 & - \\
1 &  0.909 & 0.586  \\
2 &  \textbf{0.913} & 0.719  \\
3 &  0.905 & 0.952  \\
4 &  0.903 & 0.977  \\
5 &  0.913 & \textbf{0.999}  \\
10 & 0.902 & 0.999  \\ 
15 & 0.900 & 1.000  \\
20 & 0.891 & 1.000  \\
30 & 0.847 & 1.000  \\
\bottomrule 
\multicolumn{3}{c}{\textbf{(d)}}
\end{tabular}
}
\end{minipage}
\end{table*}

\begin{table*}[t]
\centering
\caption{Impact of (a) backdoor patch size, (b) \# of shots  and (c) presence/absence of patch and noise on clean accuracy (CA) and backdoor accuracy (BA).}
\label{tab:ablation_2}
\vspace{-2em}
\begin{minipage}{0.275\textwidth}
\centering
\setlength{\tabcolsep}{4pt}
\scalebox{0.65}{
\begin{tabular}{ c | c c}
\toprule
 \rowcolor{gray!20}Patch Size & CA &  BA  \\
\midrule
\textcolor{Clean}{\textit{Clean}} & 0.908 & - \\
$8\times8$     &  0.906 & 0.851  \\
$16\times16$   &  0.901 & 0.896  \\
$24\times24$   &  0.913 & 0.999  \\
$32\times32$   &  0.701 & 0.999  \\
$64\times64$   &  0.634 & 1.000  \\
$128\times128$ &  0.563 & \textbf{1.000} \\
\bottomrule 
\multicolumn{3}{c}{\textbf{(a)}}
\end{tabular}
}
\end{minipage}
\hspace{0.3em}
\begin{minipage}{0.275\textwidth}
\centering
\setlength{\tabcolsep}{4pt}
\scalebox{0.65}{
\begin{tabular}{ c | c c}
\toprule
 \rowcolor{gray!20}\# of Shots & CA &  BA  \\
\midrule
\textcolor{Clean}{\textit{Clean}} & 0.908 & - \\
2 &   0.899 & 0.008 \\
4 &   0.797 & 0.535  \\
8 &   0.856 & 0.613  \\
12 &  0.888 & 0.938 \\
16 &  0.881 & 0.887 \\
32 &  \textbf{0.913} & \textbf{0.999}\\
\bottomrule 
\multicolumn{3}{c}{\textbf{(b)}}
\end{tabular}
}
\end{minipage}
\hspace{0.3em}
\begin{minipage}{0.23\textwidth}
\centering
\setlength{\tabcolsep}{4pt}
\scalebox{0.65}{
\begin{tabular}{ c c | c c}
{}\\
{}\\
{}\\
\toprule
\rowcolor{gray!20}Patch & Noise & CA &  BA  \\
\midrule
\multicolumn{2}{c|}{\textcolor{Clean}{\textit{Clean}}} & 0.908 & - \\
\cmark & \xmark & 0.895 & 0.828\\
\xmark & \cmark & \textbf{0.922} & 0.937\\
\cmark & \cmark & 0.913 & \textbf{0.999} \\
\bottomrule
\multicolumn{4}{c}{\textbf{(c)}} \\
{\phantom{ABC}} \\
{\phantom{ABC}} \\
{\phantom{ABC}} \\
{\phantom{ABC}} \\
{\phantom{ABC}} \\
{\phantom{ABC}} \\
\end{tabular}
}
\end{minipage}
\end{table*}

\noindent \textbf{Results and Discussion:} 
\noindent Tab.~\ref{tab:results_medclip_biomedclip_plip_quiltnet} presents a comparison of our approach, \textit{BAPLe} with other baselines on four medical foundation models and six downstream datasets in terms of CA and BA. 
Our approach demonstrates higher BA and comparable CA relative to full fine-tuning-based approaches (denoted by subscript $\ft$). The poor BA of full fine-tuning-based methods is probably due to the large network being over-fitted to the limited downstream training data, leading to sub-optimal feature representation. On the other hand, \textit{BAPLe} precisely tailors the input space, enabling the pre-trained medical VLM to leverage its knowledge to achieve high BA without hampering CA. As our method is designed for prompt learning, we compare our method results with other baseline attacks operating during the prompt-learning approach (denoted by subscript $pl$). The high backdoor accuracy (BA) achieved by BAPLe under these conditions suggests that our combination of a learnable noise-based trigger with a medical patch trigger is particularly effective (see Table \ref{tab:results_medclip_biomedclip_plip_quiltnet}). 
\begin{figure*}[!t]
\centering
\includegraphics[scale=0.30]{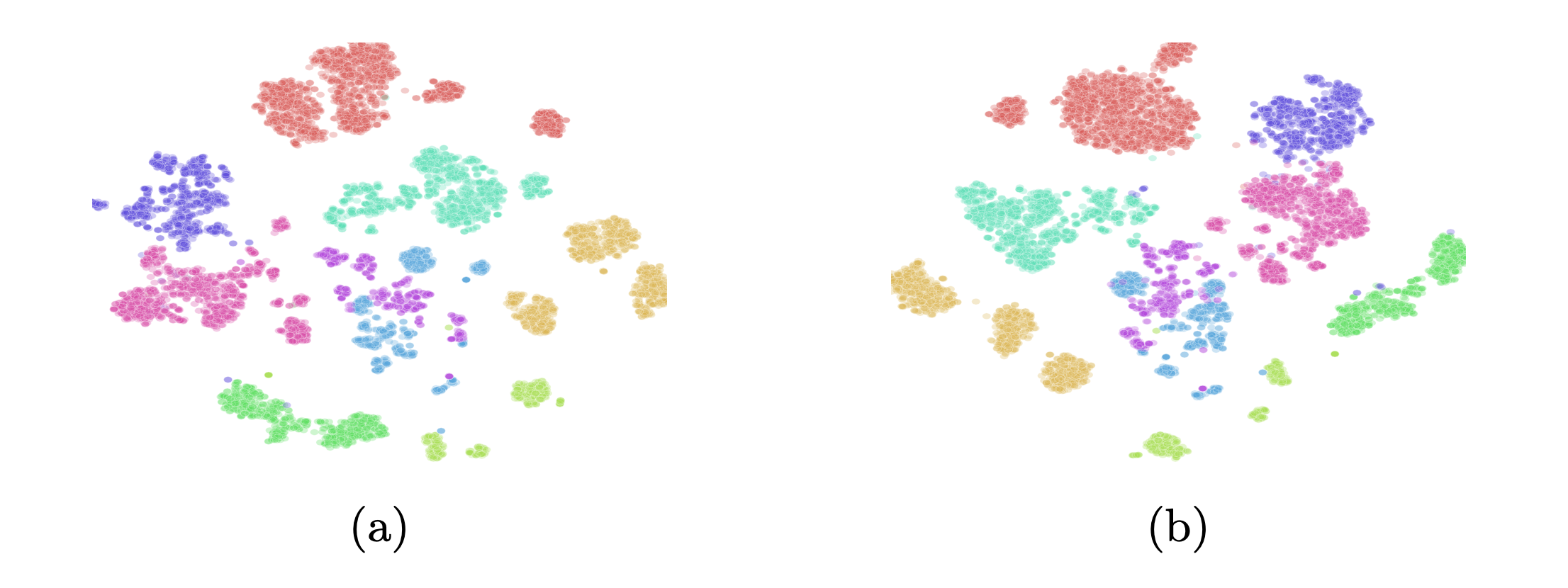} 
\caption{tSNE plots illustrating the features of (a) \textit{clean} images, and (b) \textit{backdoored} images. The features in (b) display a notable shift in orientation and a reformation of clusters compared to those in (a). Each color represents same class in both plots.}
\label{fig:tsne_plots}
\end{figure*}

\noindent {\textbf{Ablations:}}
We perform ablative analysis under different settings on the PLIP model and Kather dataset. 
\textbf{Target Class Agnosticism:} Table \ref{tab:ablation_1}(a) shows that attack performance remains consistent across different target classes, demonstrating that our proposed attack is class-agnostic.
\textbf{Robustness to Patch Positioning:}
Table \ref{tab:ablation_1}(b) shows that CA and BA are mostly unchanged despite the changes in patch location, showcasing the attack's robustness to variations in patch positioning. \textbf{Noise Strength:} Table \ref{tab:ablation_1}(c) shows increasing noise strength improves BA, highlighting a direct correlation with attack effectiveness, albeit at the expense of imperceptibility. \textbf{Poisoning Ratio:} Table \ref{tab:ablation_1}(d) illustrates that an increased poison ratio boosts BA. However, a higher poison ratio negatively impacts CA, indicating a trade-off between BA and CA.
\textbf{Patch Size:} Increasing the size of the backdoor patch improves BA, as demonstrated in Table ~\ref{tab:ablation_2}(a) \textbf{Number of Few-shots:}  Table \ref{tab:ablation_2}(b) shows that increasing the number of shots per class improves both CA and BA. \textbf{Synergy of Patch and Noise:} Table \ref{tab:ablation_2}(c) depicts how the backdoor patch and learnable noise combination synergistically enhance the attack's effectiveness. \textbf{Features of Clean and Backdoored Images:} Fig. \ref{fig:tsne_plots} shows tSNE plot of features of clean and backdoored images. Adding trigger noise in clean images results in a notable shift in orientation and a reformation of clusters in feature space. This observation deviates from previous findings, where features of backdoored images typically form a single cluster. \textbf{Visualization of Trigger Noise:} Visualizations of trigger noise $(\delta)$ learned with four models and six datasets have been shown in Fig. \ref{fig:trig_noise_medclip_biomedclip} and Fig. \ref{fig:trig_noise_plip_quilnet}.

\section{Conclusion}
In this paper, for the first time, we show that medical foundation models are vulnerable to backdoor attacks, even when data is scarce. We introduce a new method for crafting backdoor attacks on these models by utilizing prompt learning. Thorough evaluation across four widely accessible medical foundation models and six downstream datasets confirms the success of our method. Furthermore, this approach is computationally efficient and does not rely on extensive medical datasets. Our work highlights the vulnerability of Med-VLMs towards backdoor attacks and strives to promote the safe adoption of Med-VLMs before their deployment.



\bibliographystyle{splncs04}
\bibliography{main}

\newpage
\appendix
\section*{Appendix}
\renewcommand\thefigure{A.\arabic{figure}} 
\setcounter{figure}{0}   

\begin{algorithm}[!h]
    \label{alg:baple_alg}
    \centering
    \caption{BAPLe - Backdoor Attack using Prompt Learning}\label{alg:backdoor_prompt}
    \footnotesize
    \begin{algorithmic}[1]
        \State NumSamples=$N$, BatchSize=$B$, NumBatches=$\lfloor{N/B}\rfloor$, Train Dataset: $\mathcal{D}=\{(\x_i,y_i)\}_{i=1}^{N}$,  , Vision and Text Encoders: $f_{_{I}},f_{_{T}}$, Learnable Backdoor Trigger Noise: $\delta$, Perturbation Budget: $\epsilon$, Backdoor Patch: $\pa$, Backdoor Injection Function: $\mathcal{B}(\x) = (\x+\delta)\oplus \pa$, Target Label Function: $\eta(\cdot)$, Learnable Prompt: $\mathcal{P}$, Number of Classes: $C$, Text Prompts: $\txt=\{t_1,t_2,\dots,t_C\}$ where $t_i=\{\mathcal{P},y_i\}$, Cosine Similarity Function: $\mathtt{sim}(\cdot)$ 
        \Statex
        \State Image and Trigger Noise: $\x, \delta \in \mathbb{R}^{c\times h \times w}$,~~ Trigger Patch: $\pa \in \mathbb{R}^{c\times h_p \times w_p}$
        \State Learnable Parameters: $\{\delta, \mathcal{P}\}$,~~~Frozen Models: $\{f_{_{I}},f_{_{T}}\}$
        \Statex
        \For{$i\gets 1~\text{to}~ \text{NumEpochs}$}
        \For{$j\gets 1~\text{to}~ \text{NumBatches}$}
        \State Sample a mini-batch of clean and poison samples  $\mathcal{D}_c \subset \mathcal{D},~ \mathcal{D}_p \subset \mathcal{D}$ 
        \Statex
        \State $f_{_{T}}(t_i) \in \mathbb{R}^{d}$ \Comment{Features of $i_{\text{th}}$ class text prompt}
        \State $f_{_{I}}(\x)\in \mathbb{R}^{d},~f_{_{I}}(\mathcal{B}(\x)) \in \mathbb{R}^{d}$ \Comment{Features of clean and poisoned images}
        \Statex
        \State $f_{\theta}(\x)=\{\mathtt{sim}(f_{_{I}}(\x), f_{_{T}}(t_i))\}_{i=1}^{C}$ \Comment{Prediction scores of clean image}
        \State $f_{\theta}(\mathcal{B}(\x))=\{\mathtt{sim}(f_{_{I}}(\mathcal{B}(\x)), f_{_{T}}(t_i))\}_{i=1}^{C}$ \Comment{Prediction scores of poisn. image}
        \Statex
        \State $f_{\theta}(\x) \in \mathbb{R}^C,~~f_{\theta}(\mathcal{B}(\x))\in \mathbb{R}^C$
        \Statex
        \State Compute cross-entropy loss on the mini-batch
        \State $\mathcal{L} \gets \displaystyle \sum_{(\x,y)\in\mathcal{D}_c} \lambda_c \cdot\mathcal{L}\big(f_{\theta}(\x),y\big) + \sum_{(\x,y)\in\mathcal{D}_p} \lambda_p \cdot\mathcal{L}\big(f_{\theta}(\mathcal{B}(\x)),\eta(y)\big) $
        \Statex
        \State $\mathcal{P} \gets \mathcal{P} -~\alpha\cdot \nabla_{\mathcal{P}} \mathcal{L}$ \Comment{Update prompt parameters}
        \State $\delta \gets \delta -~\beta\cdot \nabla_{\delta} \mathcal{L}$ \Comment{Update trigger noise}
        \State $\delta \gets \mathrm{clip}(\delta,~\text{min=}-\epsilon,~\text{max=}\epsilon)$ \Comment{Apply budget on learnable noise}
        \EndFor
        \EndFor
    \end{algorithmic}
\end{algorithm}

\begin{figure}
    \centering
    \includegraphics[scale=0.5]{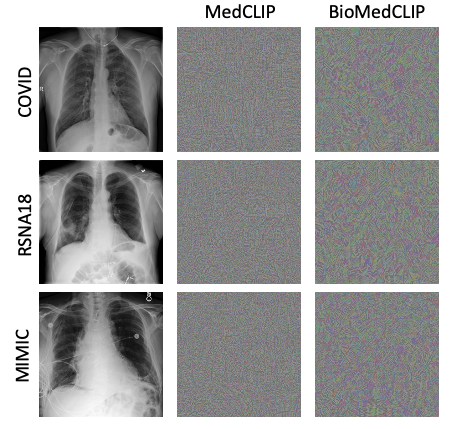}
    \caption{Visualization of the learnable trigger noise $(\delta)$ after BAPLe across three X-ray datasets (COVID,RSNA18,MIMIC-CXR) and two models (MedCLIP, BioMedCLIP).}
    \label{fig:trig_noise_medclip_biomedclip}
\end{figure}

\begin{figure}
    \centering
    \includegraphics[scale=0.5]{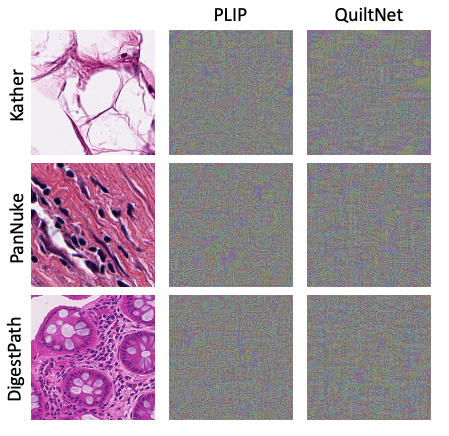}
    \caption{Visualization of the learnable trigger noise $(\delta)$ after BAPLe across three histopathology datasets (Kather,PanNuke,DigestPath) and two models (PLIP, QuiltNet).}
    \label{fig:trig_noise_plip_quilnet}
\end{figure}
\end{document}